\documentclass[letterpaper, 10 pt, conference]{ieeeconf}  

\usepackage{amsmath}
\usepackage{graphicx}
\usepackage{booktabs}
\usepackage{cite}
\usepackage{stfloats}
\usepackage{tcolorbox}
\IEEEoverridecommandlockouts                              

\title{\LARGE \bf
MindEye-OmniAssist: A Gaze-Driven LLM-Enhanced Assistive Robot System for Implicit Intention Recognition and Task Execution
}

\author{Zejia Zhang$^{1}$, Bo Yang$^{2}$, Xinxing Chen$^{1}$, Weizhuang Shi$^{1}$, Haoyuan Wang$^{1}$,\\ Wei Luo$^{3}$ and Jian Huang$^{1}$, \IEEEmembership{Senior Member, IEEE}
\thanks{*This work was supported in part by National Natural Science Foundation of China under Grant U24A20280, and in part by Hubei Provincial Technology Innovation Program under Grant 2024BAA007.}
\thanks{$^{1}$Zejia Zhang, Xinxing Chen, Weizhuang Shi, Haoyuan Wang and Jian Huang is with Hubei Key Laboratory of Brain-Inspired Intelligent Systems, Huazhong University of Science and Technology, Wuhan 430074, China, and also with Key Laboratory of the Ministry of Education for Image Processing and Intelligent Control, School of Artificial Intelligence and Automation, Huazhong University of Science and Technology, Wuhan 430074, China. \texttt{email:zejiazhang@hust.edu.cn}, \texttt{cxx@hust.edu.cn}, \texttt{why427@hust.edu.cn}, \texttt{swz@hust.edu.cn}, \texttt{huang\_jan@mail.hust.edu.cn}}%
\thanks{$^{2}$Bo Yang is with state key laboratory of intelligent vehicle safety technology, chongqing changan automobile co ltd, chongqing 400023, China. \texttt{email:ybandbob@gmail.com}}%
\thanks{$^{3}$Wei Luo is with Science and technology innovation center, China ship development and design centre, Wuhan 430060, China
\texttt{email:171772014@qq.com}}%
\thanks{Corresponding author: Xinxing Chen.}
}

\begin{document}

\maketitle
\thispagestyle{empty}
\pagestyle{empty}

\begin{abstract}
A promising effective human-robot interaction in assistive robotic systems is gaze-based control. However, current gaze-based assistive systems mainly help users with basic grasping actions, offering limited support. Moreover, the restricted intent recognition capability constrains the assistive system's ability to provide diverse assistance functions. In this paper, we propose an open implicit intention recognition framework powered by Large Language Model (LLM) and Vision Foundation Model (VFM), which can process gaze input and recognize user intents that are not confined to predefined or specific scenarios. Furthermore, we implement a gaze-driven LLM-enhanced assistive robot system (MindEye-OmniAssist) that recognizes user’s intentions through gaze and assists in completing task. To achieve this, the system utilizes open vocabulary object detector, intention recognition network and LLM to infer their full intentions. By integrating eye movement feedback and LLM, it generates action sequences to assist the user in completing tasks. Real-world experiments have been conducted for assistive tasks, and the system achieved an overall success rate of 41/55 across various undefined tasks. Preliminary results show that the proposed method holds the potential to provide a more user-friendly human-computer interaction interface and significantly enhance the versatility and effectiveness of assistive systems by supporting more complex and diverse task.

\end{abstract}

\section{INTRODUCTION}
\begin{figure*}[ht]
\centerline{\includegraphics[width=0.8\textwidth]{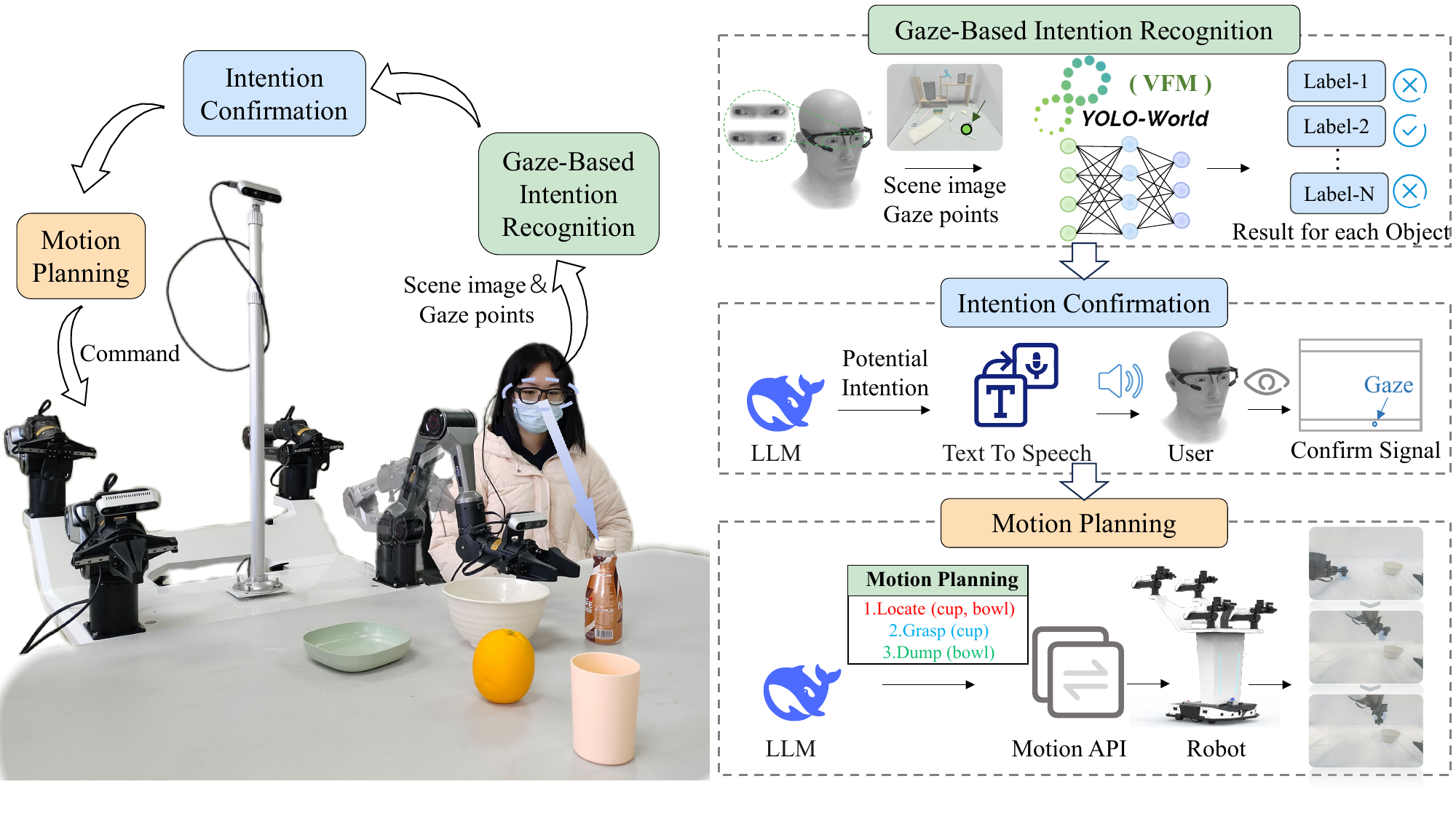}}
\caption{The schematic diagram of the MindEye-OmniAssist. }
\label{framework}
\end{figure*}

Limb function plays a crucial role in enabling individuals to perform daily tasks. However, individuals with impaired limb function may struggle to independently perform tasks\cite{impaired}. Assistive robotic systems can enhance and support the user’s limb capabilities\cite{assistive_system1}. One promising approach to facilitate effective human-robot interaction in these systems is gaze-based control\cite{HRI1,HRI2}. This approach has significant potential because the human gaze is inherently intentional and goal-directed, and individuals with limb disabilities can still maintain control over their eye movements, providing valuable spatial cues about objects with which they intend to interact. Given these advantages, gaze-based assistive systems are essential and promising.

Various gaze-based assistive systems have been developed in recent years. Cio et al.\cite{Cio} enabled the user to select the object they wish to grasp by focusing their gaze on it. Wang et al. \cite{Wang} and Li et al. \cite{Li} allowed users to control the robotic arm performing corresponding grasping tasks using gaze signals. Obviously, the above systems can only provide users with basic grasping functions. In the system developed by Wang et al.\cite{first_Wang}, the user triggers predefined tasks for an object by gazing at it. However, the predefined tasks only include picking up a cup and pouring cereal into a box.  Shafti et al. \cite{Shafti} employed gaze to recognize user's intention, but they control the action sequences using a finite state machine (FSM), which means all tasks must be predefined in the FSM. Moreover, the object recognition function is limited by the preset database, so the system cannot handle objects or tasks that are not defined. In summary, current gaze-based assistive systems primarily focus on helping users perform basic grasping actions, and the range of assistive tasks they can offer is highly limited.

Recognizing user intent is essential for the assistive system to provide effective support. To provide more auxiliary functions, the system needs to be able to recognize a wider range of user intentions. Gaze-based implicit intention recognition\cite{gao} refers to the process of analyzing a user's gaze to identify objects of interest and infer the activities the user intends to perform. For example, if a user sequentially shows interest in toothpaste, a toothbrush, and a cup, the potential intention might be brushing teeth. This method can be integrated into the gaze-based assistive system, enabling it to recognize a wider variety of intentions. There are also studies on gaze-based implicit intention recognition. Li et al.\cite{Li_intention} developed two intention knowledge bases using surveys and experimental observations, achieving accurate identification of intentions related to specific objects. Koochaki\cite{Koochaki} presents a new framework to predict the user's intention. Nevertheless, they focused only on four daily activities and twelve objects. Gao et al.\cite{gao} introduced a method combining data-driven and human prior knowledge to infer intentions, making it more consistent with human logic. However, compared to real-life situations, the intentions and object selections in the study were quite limited. Furthermore, the performance of their method largely depends on the predefined object selections in the prior knowledge matrix, making it difficult to apply to other scenarios and limiting its generalization ability. As discussed above, current research on implicit intention recognition relies heavily on artificially constructed domain-specific prior knowledge, limiting the range of intentions that can be recognized. This restricted intent recognition capability further constrains the assistive system's ability to provide diverse assistance functions.

In response to the above issues, this study aims to introduce foundation models to develop a gaze-based assistive system which is capable of recognizing intentions beyond predefined scenarios and providing auxiliary functions. Large Language Models (LLMs) \cite{LLM1,LLM2,LLM3} and Vision Foundation Models (VFMs) \cite{VLM1,VLM2,VLM3} are referred to as foundation models due to their scale and the generality of their training data, which enable downstream applications to address a wide range of tasks. However, LLMs rely heavily on textual input (such as user queries and prompts) to generate appropriate responses. Therefore, it is still difficult for LLMs to operate effectively in gaze-based scenarios. We have designed a framework that integrates visual cues with foundation models to enhance scene perception, intention reasoning, and action planning capabilities of the proposed system. The main contributions of this study are as follows.

(1) A gaze-based open implicit intention recognition framework powered by LLM and VFM. This framework is capable of processing gaze input, and addressing undefined user intentions.

(2) A gaze-driven LLM-enhanced assistive robot system (MindEye-OmniAssist) that recognizes user’s implicit intentions through gaze and plans action sequences to assist in completing task. Compared to current systems, it is capable of performing a wider range of assistive tasks.

Real-world experiments have been conducted to verify the effectiveness of the proposed MindEye-OmniAssist system. The system achieved an overall success rate of 41/55 across various undefined tasks, with particularly high recognition (52/55) and planning (50/52) success rates. Our system can inspire future work on assistive systems using gaze as a human-machine interaction interface, encouraging further exploration of how to recognize complex user intentions and adapt to a broader set of tasks.


\section{System Overview}

\subsection{Hardware Architecture}
The hardware setup of the MindEye-OmniAssist consists of a Pupil Invisible head-mounted eye tracker, four collaborative robotic arms, three Intel RealSense D435 series cameras and a laptop
. The Pupil Invisible’s scene camera is capable of outputting scene images at a resolution of $1088\times 1080$ and a rate of 30 Hz. Furthermore, the eye tracker provides realtime 2D gaze points $[g_x, g_y]$ on the scene image, at a frequency of about 120 Hz. The robot has four arms, comprising two main arms and two follower arms, offering two operating modes: in teleoperation mode, the follower arms move in tandem with the main arms; in autonomous mode, the follower arms move based on code instructions, while the main arms remain stationary. The three cameras provide RGB and depth images with different fields of view. All devices communicate with a PC running Ubuntu operating system. The assistive robotic system implemented based on Robotic Operation System (ROS).

\begin{figure*}[bp]
\centerline{\includegraphics[width=0.8\textwidth]{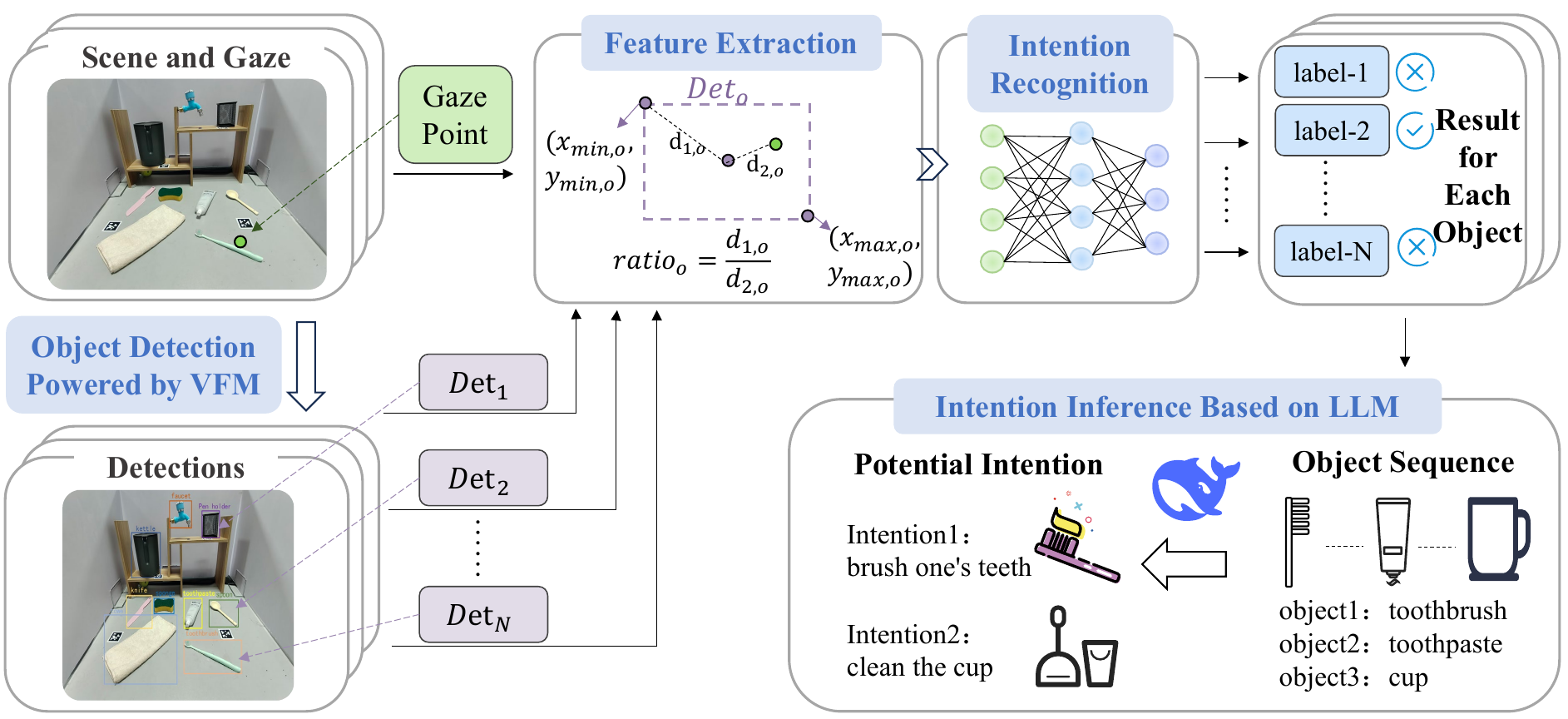}}
\caption{Open implicit intention recognition framework based on gaze and foundation models. }
\label{intention framework}
\end{figure*}
\subsection{Software Architecture}
The architecture of the MindEye-OmniAssist, depicted in Fig.\ref{framework},  includes a gaze-based implicit intention recognition module, a user intention confirmation module and a motion planning module. Firstly, the system utilizes the gaze-based intention recognition module to recognize the user’s intentions based on natural gaze data and scene images. Once the intention is recognized, the system will verify it with the user. After confirmation, it proceeds to plan the required action and calls the corresponding custom-developed Application Programming Interface (API) to execute the task. In the following section, we describe the details of the three modules.

\section{Method}
\subsection{Gaze-based Intention Recognition } 
To achieve open implicit intention recognition, we designed an intention recognition framework based on gaze and foundation models, as shown in Fig.\ref{intention framework}. It combines an open vocabulary object detector, a high-accuracy intention recognition network, and a LLM. The open vocabulary object detector is initially employed to identify objects in scene images. Subsequently, the intention recognition network classifies each object based on whether the user has an interaction intention with it, effectively performing a binary classification for each object. Finally, all objects that the user intends to interact with are input into the LLM to infer the user's complete intentions, such as pouring water into the bowl or fetching the bottle. In what follows, we describe the details of the framework.

\textbf{Object Detection Powered by VFM}. We use the open vocabulary object detector YOLO-World model \cite{YOLO} to identify objects in scene images and record the recognition results. For each object $o$ in the scene image, the recognition result is represented as:
\begin{equation}\label{detection_result}
Det_o = \{ x_{\text{min},o}, y_{\text{min},o}, x_{\text{max},o}, y_{\text{max},o} \},
\end{equation}
where $x_{\text{min},o}$ and $y_{\text{min},o}$ are the coordinates of the bottom-left corner of the object's bounding box, and $x_{\text{max},o}$ and $y_{\text{max},o}$ are the coordinates of the top-right corner of the bounding box, defining the object's spatial extent in the image.

\begin{figure}[!t]
\centerline{\includegraphics[width=0.9\columnwidth]{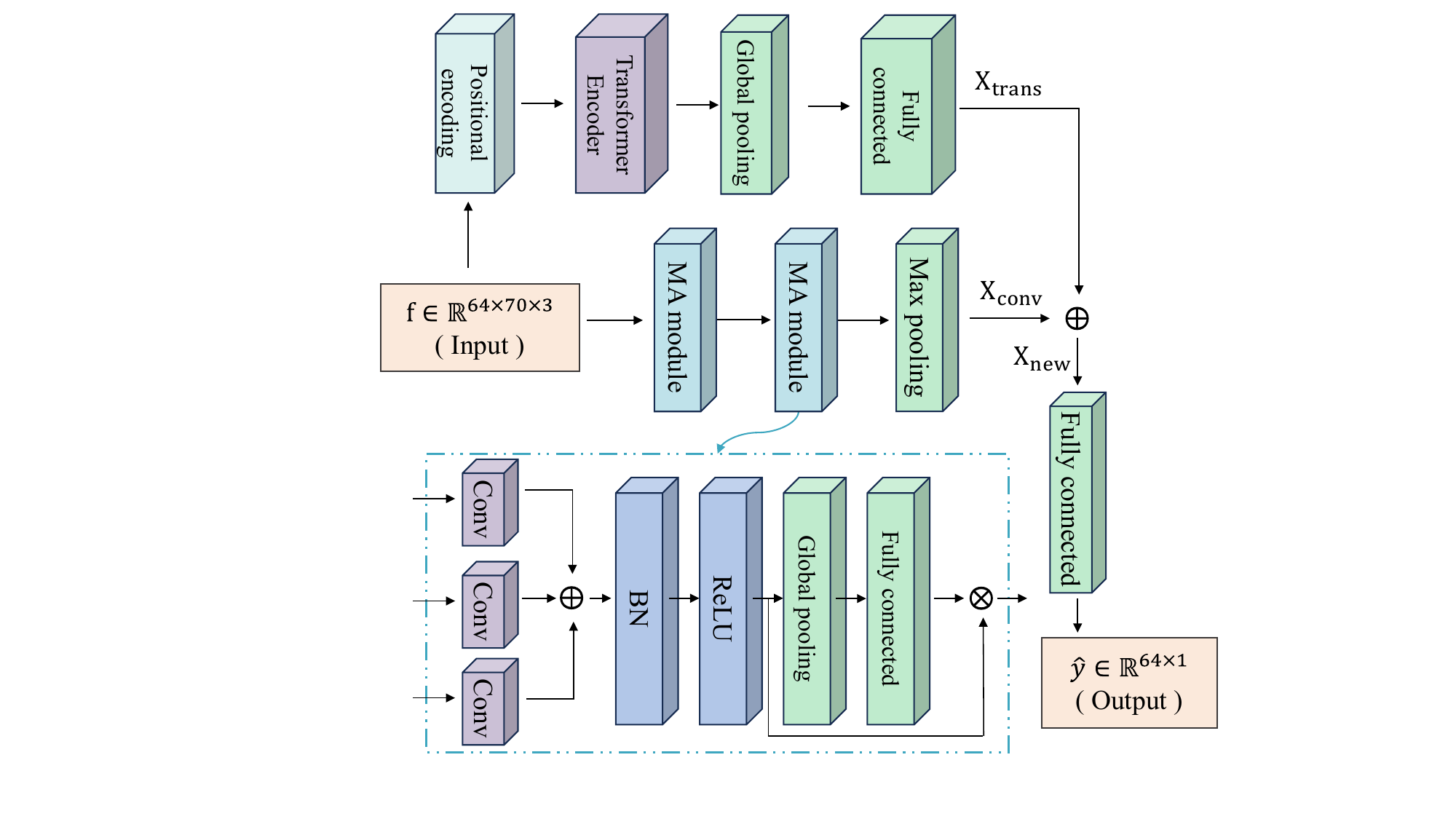}}
\caption{Architecture of the proposed intent recognition network. The network comprises multi-scale convolutional layers, positional encoders, Transformer encoders, and fully connected layers as its core components.}
\label{Network}
\end{figure}
\textbf{Feature Extraction}. For each object $o$, once its bounding box coordinates are obtained, we calculate the distance from the center point to the vertex $d_{1,o}$ and the distance from the center point to the gaze point $d_{2,o}$. $d_{1,o}$ provides information about the size of the object, $d_{2,o}$ indicates the spatial relationship between the gaze point and the object's center. The ratio of these two distances $ratio_o$ provide information about the relative position of the gaze point to the object. 
\begin{equation}\label{ratio}
ratio_o = \frac{d_{1,o}}{d_{2,o}}.
\end{equation}
The raw data consists of long data sequences obtained from each experiment. A sliding window is used to divide the sequence into multiple fixed-length segments, each with a length of $sw$. Within each segment, feature vectors are extracted:
\begin{equation}\label{feature}
 features = [g_x, g_y, ratio_o]. 
\end{equation}
The final network input $f$ is as follows:
\begin{equation}\label{input}
f \in {R}^{bs \times sw \times num\_fea},
\end{equation}
where $bs$ represents the number of samples input at each time, $sw$ is the length of the sliding window, and $num\_fea$ is the length of the feature vector, which in this paper is three.

\textbf{Intention Recognition}. The network consists of several key modules, including multi-scale convolutional layers, positional encoders, Transformer encoders, and fully connected layers. The input data $f$ is processed through three multi-scale convolutional layers with kernels of scales (3,7,13) to extract features from varying receptive fields \cite{MA}. Subsequently, the outputs are concatenated along the channel dimension to form a comprehensive feature representation. To further improve feature selection ability, attention mechanism is introduced. The resulting convolutional features are denoted as $X_{\text{conv}}$.

To capture positional relationships within time-series data, we also introduce positional encoding to the input data $f$. The features after positional encoding are fed into a Transformer encoder \cite{Transformer} to model the global dependencies among the features. After the feature vector $X_{\text{trans}}$ is obtained, the information from both branches is then merged and mapped to a new feature space through a fully connected layer, and a Sigmoid activation function is used to produce the final classification result $\hat{y}$. If $\hat{y}$ is greater than 0.5, it means there is an intention to interact with the object; otherwise, there is no intention to interact.

\textbf{Intention Inference Based on LLM}. Implicit intention recognition requires inferring the user's complete intentions based on identifying which objects they intend to interact with. Models trained on large corpora possess a robust knowledge base and inferential capabilities. Utilizing language models to understand common object combinations and their associated activities allows for an effective integration of low-level object recognition with high-level behavioral inference. We leverage the knowledge base of DeepSeek-R1 \cite{deepseek} to deduce users' intentions in various scenarios
.


The command from the user is turned into a prompt suitable for the LLM. With this prompt as an input, the LLM outputs the potential intention. Specifically, the object labels $\langle object1\rangle$, $\langle object2\rangle$ (and possibly one or more others) provided by the user is transferred into a LLM prompt in the following format:

\begin{tcolorbox}[colback=gray!15, 
                 arc=3mm, 
                 boxrule=0pt, 
                 left=2mm, right=2mm, top=2mm, bottom=2mm, 
                 boxsep=2pt 
                ]  
\textit{
``role": ``system", ``content":
} ``You are a personal assistant who infers what the user wants to do based on the objects they are looking at." \\
\textit{
``role": ``user", ``content":
} ``When the user looks at $\langle object1\rangle$, $\langle object2\rangle$, etc., in sequence, what are the possible intended actions? Please provide up to three possible intentions."
\end{tcolorbox}

\subsection{Intention Confirmation}
\begin{figure}[!b]
\centerline{\includegraphics[width=0.4\columnwidth]{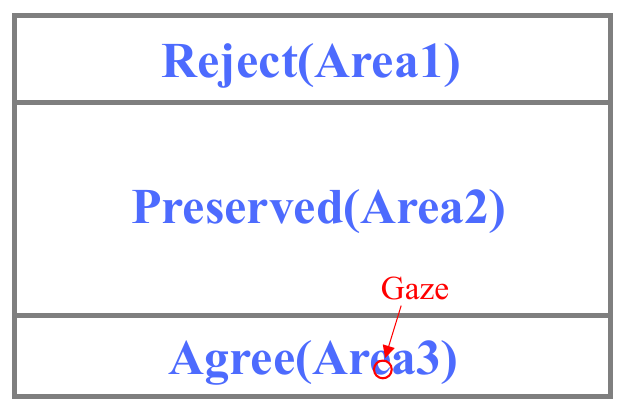}}
\caption{Diagram of the user's confirmation using the gaze command. When the user's gaze point falls on Area 3, it indicates agreement.}
\label{confirm}
\end{figure}

To confirm with the user whether the inferred intention is correct, we designed an intention confirmation module. The $1088 \times 1080$ scene images captured by the eye tracker’s scene camera are used as the control interface. The image is divided into three sections, as shown in Fig.\ref{confirm}. In normal circumstances, the user's gaze is centered in the middle area, so no instructions are assigned to the central region (Area 2), leaving it for normal observation behavior during the interaction. The upper region (Area 1) and lower region (Area 3) are designated for issuing ``Reject" and ``Agree" commands, respectively. The user can issue the corresponding command by shifting their gaze to the appropriate area.

Once the user's intention $\langle intention\rangle$ is inferred, the system will confirm it with the user via voice: ``Is your intention $\langle intention\rangle$?". Upon hearing this confirmation, if the inferred intention is correct, the user is expected to look at Area 3 to indicate agreement. Conversely, if the intention is incorrect, the user should look at Area 1 to reject the execution of the inferred intention. 

\subsection{Motion Planning}
After confirming the user's intention, the system needs to calculate the sequence of robotic arm actions to assist in completing the task. In this study, following the approach of Liang et al.\cite{cap}, we utilize the LLM to generate action sequences based on user intentions. To achieve this, we define a set of operation APIs, which cover basic actions such as object localization, grasping, placing, moving, and pouring. The simple operations are directly pre-programmed, while the complex operations are learned in advance using a behavior cloning algorithm.

\section{Experiments and Results}
\subsection{Experimental Setup}
\textbf{1) Intention Recognition Evaluation:} To validate the performance of the proposed single object intention recognition network, experiment-based performance evaluation and subject-based performance evaluation are conducted. In the experiment-based evaluation, five-fold cross-validation is used to prevent overfitting from similar eye movement behavior within trials, with one trial as the test set and the others for training in each iteration.  In the subject-based evaluation, the performance of the algorithm is assessed across different subjects by using the data of one subject as the test set and others for training set, repeating for each subject. The proposed algorithm is compared with Midas-Net\cite{MIDAS} and gaze-based methods \cite{Shafti}, using accuracy as the evaluation metric. Accuracy is calculated by comparing the model's predicted intention ($\hat{y}$) with the actual intention label ($y$). 


\textbf{Dataset}. To complete the above evaluation, we need a dataset where users express implicit intentions through eye movements. Since there is no such open-source dataset available at present, we established an eye movement dataset using the Pupil Invisible eye tracker. The dataset includes data from eight volunteers, aged 20 to 24, and contains various objects, including bottles, scissors, bananas, grapes, etc. The experimental setup involved subjects sitting in front of a table with these objects, wearing the eye tracker, and performing tasks designed to express interaction intentions through gaze
. 
In one part of the experiment, participants expressed their intention to interact with objects, such as gazing at a banana to signal picking it up, or gazing at a kettle and cup to indicate pouring water. In another part, participants gazed unconsciously to simulate natural eye movement. The position of objects was changed throughout the experiment and each task was repeated five times. Eye tracker calibration was performed before each round to ensure accurate gaze estimation, and recalibration occurred if accuracy dropped during the experiment. Approximately 120 minutes of data were collected across all trials.

After data collection, several post-processing steps were applied to align the data along the time axis and mitigate the influence of head movements on gaze points. Additionally, the intentions behind participants' eye movements were manually annotated to create labeled data for training and evaluation. Therefore, data at each moment in the dataset includes a scene image, 2D gaze point coordinates relative to the scene image, and manually annotated labels indicating the presence of interaction intent.

\textbf{2) System Evaluation:}
We tested our system in a desktop environment. Five participants took part in the experiment.Subjects wore the eye-tracker and were seated approximately 35 cm away from the table. The eye-tracker was calibrated prior to the start of each trial to ensure accurate gaze tracking. If any drift or inaccuracy was observed during the experiment, recalibration was performed to maintain data precision. During the experiment, the participants' gaze was not disturbed, allowing for natural behaviors, such as blinking. 

\begin{table}[!t]
\centering
\caption{Experiment-based performance evaluation results}
\begin{tabular}{cc}
\toprule
\textbf{Algorithm} & \textbf{Accuracy} \\
\midrule
\textbf{Ours} & \textbf{96.52 $\pm$ 1.69\%} \\
Midas-Net \cite{MIDAS} & 91.13 $\pm$ 2.44\% \\
Method based on fixation \cite{Shafti} & 61.17  $\pm$ 2.93\% \\
\bottomrule
\end{tabular}
\label{tab:Experiment-based performance}
\end{table}

\begin{table*}[!t]
\centering
\caption{System Evaluation Experimental Results}
\label{system result}
\begin{tabular}{ccccc}
\toprule
                             & Overall(S/ALL) & Recognition(S/ALL) & Plan(S/ALL) & Execution(S/ALL) \\ \midrule
Fetch an Object              & 16/20          & 20/20              & 20/20       & 16/20            \\
Put Something into Something & 16/20          & 18/20              & 17/18       & 15/17            \\
Water the Plants             & 3/5            & 5/5                & 5/5         & 3/5              \\
Toggle the Switch            & 4/5            & 4/5                & 3/4         & 3/3              \\
Pour Water                   & 4/5            & 5/5                & 5/5         & 4/5              \\ 
\textbf{Total}   &\textbf{41/55}    & \textbf{52/55}   & \textbf{50/52}  & \textbf{41/50}\\ 
\bottomrule
\end{tabular}
\end{table*}

During each trial, 1 to 5 different objects were randomly placed on the table. These objects included items such as bottles, cups, bowls, etc. The participants were not given specific instructions regarding the tasks; instead, they were free to express their intentions through gaze, and the robotic arm would assist in executing the tasks accordingly. A trial was considered successful if the task was completed within three attempts. For each trial, the success rate at different stages is recorded at the same time. The performance of the system was assessed in terms of the overall success rate, the success rate of intention recognition, the success rate of action planning and the success rate of action execution. Throughout the experiment, we recorded the reason for any failures. 

\subsection{Experimental Results and Discussion}
\textbf{1) Intention Recognition Results:}
In the experiment-based performance evaluation, the average accuracies of the proposed algorithm and the comparison algorithms are shown in the Table \ref{tab:Experiment-based performance}. It is evident from the table that the accuracy of the gaze-based method is significantly lower than that of the proposed algorithm. This occurs because, instead of deliberately extending their gaze times, the subjects in the experiment maintained natural eye movements such as blinking, which could influence the accuracy of gaze detection.

\begin{figure}[!t]
\centerline{\includegraphics[width=0.8\columnwidth]{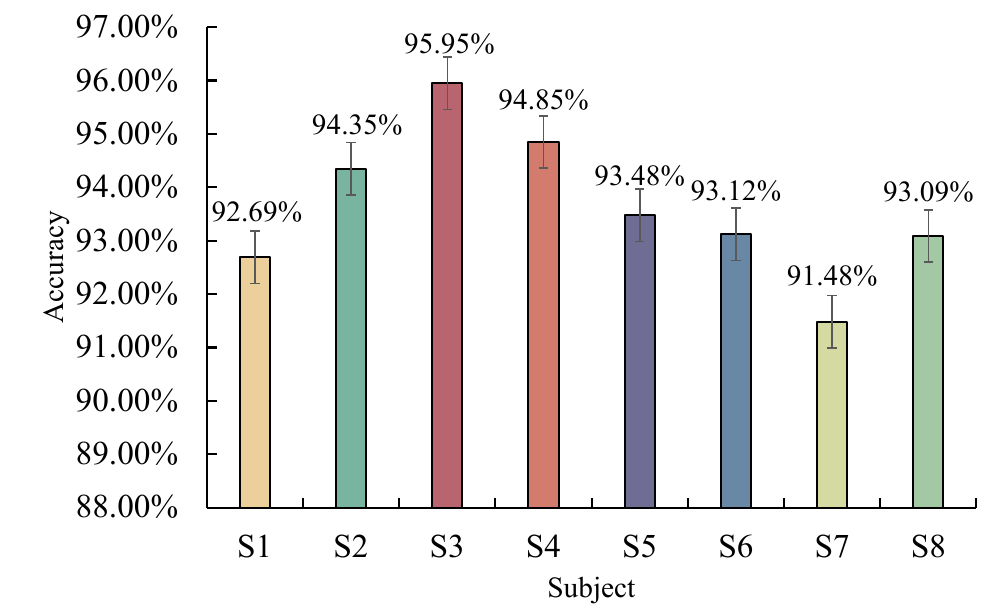}}
\caption{Test accuracy of different subjects using our method.}
\label{result}
\end{figure}

In the subject-based performance evaluation, the average testing accuracy for the eight participants was 93.63 $\pm$ 1.39\%. The testing accuracy for each participant is shown in Fig.\ref{result}. Due to variations in eye movement behavior among users, there is some variation in accuracy across participants, but with the highest accuracy reaching 95.95\% and the lowest at 91.48\%, the model demonstrates superior performance in cross-subject recognition tasks. In contrast, the average accuracies of Midas-Net and the gaze-based method are 88.18 $\pm$ 2.07\% and 59.83 $\pm$ 0.69\%, respectively.

\textbf{2) System Evaluation Results:}
The results are summarized in Table \ref{system result}. `Overall’ represents the
overall success rate of MindEye-OmniAssist, i.e., the proportion of successfully completed tasks out of all tasks. `Recognition', `Plan', and `Execution' represent the success rate of intention recognition, action planning and action execution, respectively. Each stage's success rate is calculated under the condition that the previous stage was successful. Each value is in the format ``S/ALL", where S represents the number of successful trials and ALL represents the total number of trials conducted. For example, in the ``Overall" column for the task ``Fetch an Object", the value is 16/20, indicating that the system successfully completed the task 16 times out of 20 trials.

\section{CONCLUSIONS}
Gaze-based control in assistive robotic systems holds promise for enhancing upper limb capabilities, but the limited intent recognition restricts current systems to basic tasks. In this study, we proposes a gaze-based open implicit intention recognition framework based on LLM and VFM, capable of recognizing user intentions beyond predefined scenarios. Specifically, an open-vocabulary object detector is used to identify objects in the scene, and an intention recognition network classifies each object based on whether it is a target for interaction. Interaction targets are then input into the LLM to infer the user's complete intention. In addition, we present a gaze-driven assistive robotic system. After inferring the intention through gaze signals, the system provides the intention to the user through voice. The user provides corrective feedback through eye movement signals, and the system subsequently generates customized action sequences based on the LLM to assist the user in completing tasks. The experimental results indicate that our system can effectively identify user intentions and assist users in completing various tasks. Future work will focus on improving the accuracy of action planning , developing more effective methods for updating the basic action library, and implementing error correction mechanisms.

\addtolength{\textheight}{-1cm}   





\bibliographystyle{IEEEtran}
\bibliography{IEEEabrv,myref_new}

\end{document}